\documentclass[runningheads]{llncs}

\usepackage[T1]{fontenc}

\usepackage{graphicx}
\usepackage{comment}
\usepackage{amsmath,amssymb}
\usepackage{color}
\usepackage{url}
\usepackage{hyperref}

\newif\ifreview
\reviewfalse

\ifreview
	\usepackage{lineno}

	\linenumbers
\fi

\begin{document}


\def\SubNumber{086}

\def\GCPRTrack{Fast Review Track}

\title{Automatic Reverse Engineering: Creating computer-aided design (CAD) models from multi-view images}

\ifreview
\else
	\titlerunning{Automatic Reverse Engineering}

	\author{Henrik Jobczyk
	\and
	Hanno Homann
	}
	
	\authorrunning{H.~Jobczyk and H.~Homann}
	
	\institute{Hannover University of Applied Sciences, Germany \email{henrik.jobczyk@stud.hs-hannover.de, hanno.homann@hs-hannover.de}}

\fi

\maketitle              

\begin{abstract}
Generation of computer-aided design (CAD) models from multi-view images may be useful in many practical applications. To date, this problem is usually solved with an intermediate point-cloud reconstruction and involves manual work to create the final CAD models. In this contribution, we present a novel network for an automated reverse engineering task. Our network architecture combines three distinct stages: A convolutional neural network as the encoder stage, a multi-view pooling stage and a transformer-based CAD sequence generator. 

The model is trained and evaluated on a large number of simulated input images and extensive optimization of model architectures and hyper-parameters is performed. A proof-of-concept is demonstrated by successfully reconstructing a number of valid CAD models from simulated test image data. Various accuracy metrics are calculated and compared to a state-of-the-art point-based network.

Finally, a real world test is conducted supplying the network with actual photographs of two three-dimensional test objects. It is shown that some of the capabilities of our network can be transferred to this domain, even though the training exclusively incorporates purely synthetic training data. However to date, the feasible model complexity is still limited to basic shapes.

\end{abstract}
\section{Introduction}
\label{sec:intro}

Ever since the invention of 3D-printing in the middle of the 20th century, it stimulates the imagination of laypersons and engineers alike. Nowadays this technology is an integral part of the product development cycle in many industries and its application often goes beyond the production of mere prototypes.

Even though online 3D printing services increase availability at affordable prices, their use in everyday life is not straightforward. This work is focuses on the central problem of 3D-printing: The generation of digital 3D objects is a skill requiring specialized technical expertise and training, posing a significant barrier for consumer adoption.

To give a practical example, a simple mechanical part within a bigger and more expensive appliance such as a washing machine or dryer fails and renders the device unusable. The point of failure is identified but the manufacturer can not offer a spare part. If the user could simply take a few photos using a smartphone camera and have a computer-aided design (CAD) model created automatically by software, the problem could be solved in a short time at minimal financial and environmental cost.

\begin{figure}[htb]
  \centering
   \includegraphics[width=1.0\linewidth]{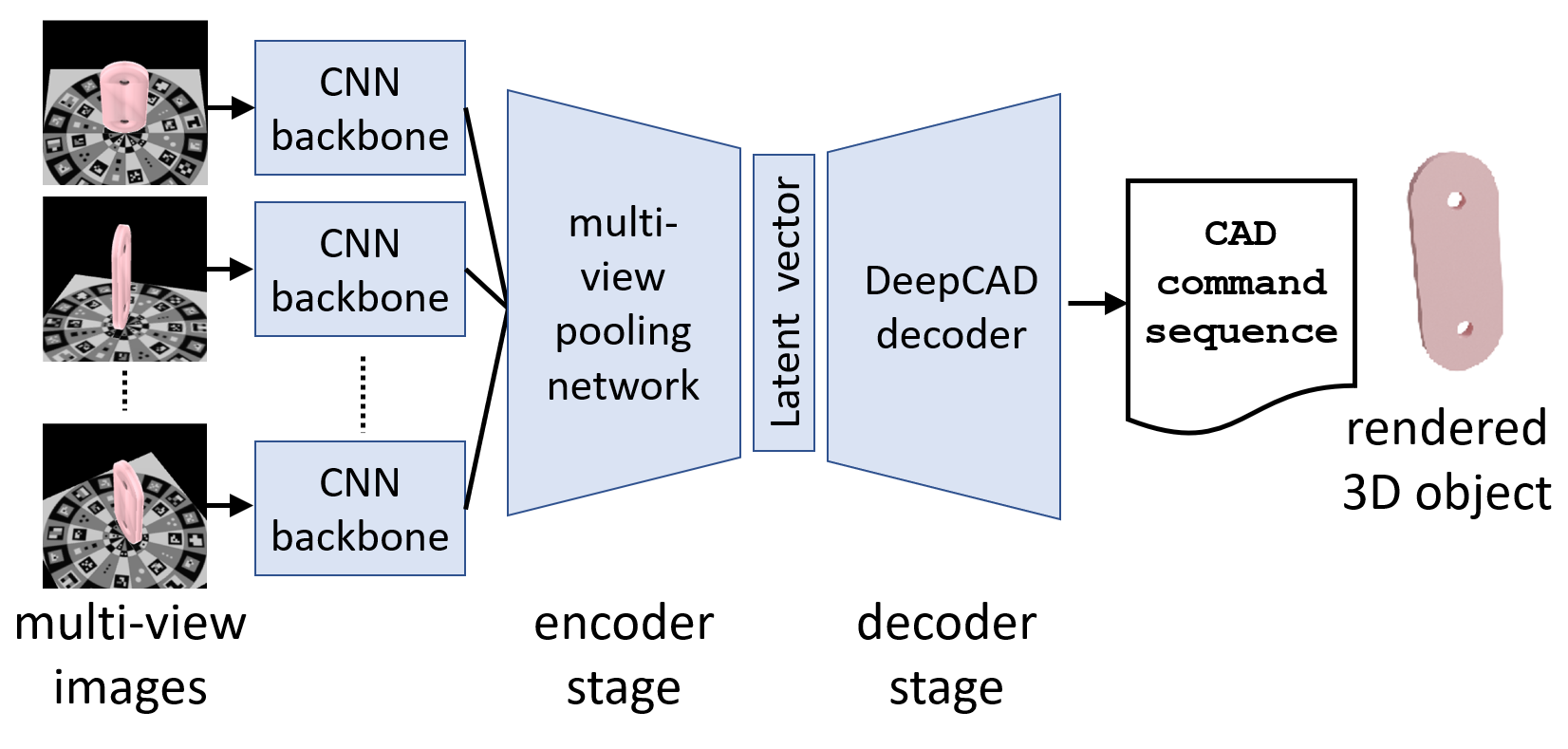}
   \caption{ARE-Net architecture: Input images taken from multiple view angles are fed into an encoder-decoder network to generate CAD sequence file. Multi-view fusion is facilitated (a) using a fully-connected network (FCN) or (b) using a gated recurrent unit (GRU) to allow varying numbers of input images. The decoder part of the DeepCAD auto-encoder is employed as the generative decoder.}
   \label{fig:mv_network}
\end{figure}

This work proposes an end-to-end solution for this reverse engineering problem, which is to our knowledge the first of its kind. Our network architecture is illustrated in \autoref{fig:mv_network} and will be described in detail further below after revisiting the state-of-the-art.
For proof-of-concept, our model was trained on a large number of renderings from simulated CAD objects. Our results indicate that the image-based approach may outperform a current point-based method. Finally, two real world objects were photographed and reconstructed.

Our main contributions are: (1) We present the first end-to-end model to generate CAD sequences from multi-view images, (2) comparison of two different multi-view fusion strategies, and (3) initial results on real-world photos.

\section{Related work}
\label{sec:state_of_the_art}
\subsection{Traditional photogrammetry approaches to reconstructing CAD models}
\label{ssec:conventional}

Photogrammetry is frequently deployed as an image-based technique to measure three-dimensional shapes using inexpensive cameras. The most common monocular approaches are based on the Structure from Motion (SfM) method first described in \cite{shimon_ullman_interpretation_1979}. Here, the software is provided with several images from different perspectives and then computes a point-cloud of the object of interest.

Automatically extracting a CAD model from a point-cloud is however not straight-forward. For example, the professional AutoCAD software can import but not post-process point clouds as of today \cite{Autodesk}. Thus far, CAD model creation mostly remains a manual task.

Kim et al.~\cite{kim2011fully} proposed 3D registration of given CAD model using the iterative closest point (ICP) method. 
Budroni et al.~\cite{budroni2010automated} have demonstrated the fitting of planar surfaces to point clouds for reconstructing of 3D models of interior rooms.  
More recently, Lui~\cite{liu2020adaptive} proposed automatic reverse-engineering of CAD models from points clouds by iteratively fitting primitive models based on the RANSAC algorithm.
In conclusion, there are few existing approaches which are however domain-specific. Instead, a neural-network based approach might generalize better in the long term.  

\subsection{Learning-based object reconstruction} 
%

Detection of 3D objects from multiple view perspectives has been addressed by Rukho\-vich et al.~\cite{rukhovich_imvoxelnet_2021}. Similar to \cite{wang_fcos3d_2021}, they used a fully convolutional network. Notably the number of monocular images in their multi-view input can vary from inference to inference, offering high versatility. This is achieved by extracting features with conventional a Convolutional Neural Network (CNN), followed by pooling and back-projecting into a 3D volumetric space. In this space, bounding boxes are predicted by three-dimensional convolutions.

For 3D surface reconstruction, deep learning models have been suggested for different kinds of object representations, including 
point clouds \cite{achlioptas2018learning,insafutdinov2018unsupervised,fan_point_2017,yang_pointflow_2019,yang_foldingnet_2018,vedaldi_learning_2020,mo_structurenet_2019}, 
triangle meshes \cite{ferrari_pixel2mesh_2018,groueix_papier-mache_2018,nash_polygen_2020,pan2019deep},
voxel grids \cite{liao_deep_2018,wu_learning_2017,yan2016perspective}, 
cubic blocks \cite{yagubbayli_legoformer_2021},
parametric surfaces \cite{sharma_parsenet_2020,wang_pie-net_2020,jayaraman_uv-net_2021,lambourne_brepnet_2021,li_sketch2cad_2020,xu_inferring_2021},
and signed distance fields (SDFs) \cite{park2019deepsdf,jiang2020sdfdiff}. 
The majority of the studies above (e.g.~\cite{achlioptas2018learning,pan2019deep,park2019deepsdf,jiang2020sdfdiff}) use auto-encoders, with a feature bottleneck between an encoder and a decoder stage. This network architecture also allows to simplify training by separating the two stages.
To date, discrete CAD models have not been investigated for 3D surface representation.

\subsection{Multi-view convolutional networks} 
\label{ssec:multi-view}
A 3D multi-view CNN (MVCNN) for object classification was introduced by Su et al.~\cite{su_multi-view_2015}. They provide a certain number of images taken of the object as input to a common CNN and pool the extracted features using an element-wise maximum operation. The pooled information is then processed by a second CNN and a final prediction is made. Notably they conclude that inputting 12 evenly spaced perspectives offers the best trade-off between prediction accuracy and memory as well as time resources.

Their concept as been studied 
for classifying 3D shapes from point clouds \cite{mohammadi_pointview-gcn_2021,qi_volumetric_2016}.
In general, working with MVCNNs seems to be a viable approach for extracting information from 3D scenes. Leal et al.~\cite{leal-taixe_deeper_2019} compared different 3D shape classifiers, identifying MVCNNs as superior to other methods due to better generalizability and outperforming several point-based and voxel-based approaches. Consequently, this approach will be followed in this work.

\subsection{Recurrent convolutional networks} 
\label{ssec:recurrent}

While  MVCNNs showed good results for classification tasks, the simple pooling methods (e.g.~element-wise max-pooling \cite{su_multi-view_2015}) might allow a single view to overrule all other views. Geometric information not visible in some  images might be lost for a 3D reconstruction task. Hence, we alternatively consider Recurrent CNNs as a more preservative information extractor.


Zreik et al.~\cite{zreik_recurrent_2019} used a Recurrent Neural Network (RNN) for spacial aggregation of extracted information from 3D angiography images after pre-processing by a 3D-CNN.
Liu et al.~\cite{liu_convolutional_2018} combined a traditional 2D CNN backbone and an RNN to synthesize multi-view features for a prediction of plant classes and conditions. After extensive experiments, they conclude that a combination of MobileNet as a backbone and a Gated Recurrent Unit (GRU) delivers the best trade-off of classification accuracy and computational overhead. Hence, GRUs will be evaluated in this study for multi-view pooling.


\subsection{Generation of CAD representations}
\label{ssec:cad_generation}

Even though most methods described above generate three-dimensional data, none of them directly attempts CAD file generation by means of generating a construction sequence comparable to manual CAD design. Thus their resulting models cannot easily be modified by an average user. However, recent research has started to address the direct generation of parametric 2D CAD models:

Willis et al.~\cite{willis_engineering_2021} first proposed generative models for CAD sketches, producing curve primitives and explicitly considering topology. 
SketchGen \cite{para_sketchgen_2021} generates CAD sketches in a  graph representation, with nodes representing shape primitives and edges embodying the constraints. 
Similarly, Ganin et al.~\cite{ganin_computer-aided_2021}  utilized off-the-shelf data serialization protocols to embed construction sequences parsed from the online CAD editor Onshape \cite{onshape_onshape_nodate}.

DeepCAD by Wu et al.~\cite{wu_deepcad_2021} was the first approach going beyond the 2D domain of CAD sketch generation. 
They formulated CAD modeling as a generation of command sequences,  specifically tailored as an input to a transformer-based auto-encoder. The publicly available Onshape API was used to build a large dataset of 3D object models for training. 

Each object is represented by a CAD sequence, consisting of three common types of commands: (1) Creation of a closed curve profile ("sketch") on a 2D plane, (2) 3D extrusions of such sketches and (3) boolean operations between the resulting 3D objects. 
Each of the CAD commands supports a number of parameters, which may be a mixture of continuous and discrete values. To conform with their neural network, Wu et al.~sort each command's parameters into a generalized parameter vector and all continuous parameters are quantized to 8-bits.
The maximum number of commands in a given CAD construction sequence was limited to 60, corresponding to the longest sequence length in the dataset.


These CAD sequences are processed by an auto-encoder, trained to compress a given CAD model into a latent vector (dimension of 256) and then to reconstruct the original model from that embedding. This means, a random but valid CAD object can be constructed using a given 256-dimensional latent vector. In this work, chose the decoder part of DeepCAD as the generative stage of our new model as introduced next.

\section{Methods}

\subsection{Network architecture}

We introduce a novel network architecture for end-to-end generation of CAD models from multiple input images. The network is composed of three stages: (1) a CNN encoder backbone to extract information from each input image individually, (2) a pooling network that aggregates this information into a common latent vector, and (3) a generative decoder network constructing the output CAD sequences. This network structure is illustrated in \autoref{fig:mv_network}.


Considering its successful track record in object detection and classification as well as its small size, we chose the residual network architecture (ResNet) \cite{he_deep_2016} as our encoder backbone. As the visual complexity of our input images is relatively low, we  assumed that a smaller, more shallow variant of the network should suffice. Thus only its smallest two variants were evaluated, namely ResNet-18 and ResNet-34. The input image size is adjustable by means of ResNet's adaptive average pooling layer. In this work, we used 128x128 monochrome as well as 224x224 RGB input images. The output of the last fully connected layer, a vector of fixed length 512, is fed into the pooling network. All input views are processed by the backbone network individually but share the same parameters. 

The task of the multi-view pooling stage is to combine the information from multiple views. We evaluated two different network architectures during the experiments: (a) a simple feed-forward fully connected network (FCN) as a baseline model and (b) a gated recurrent unit (GRU). Following \cite{cho_learning_2014} and \cite{liu_convolutional_2018}, we assume that a recurrent pooling approach should perform favorable, even though its training is inherently more challenging \cite{pascanu_difficulty_2013} because of the possible vanishing and exploding gradient problems.

The FCN pooling network concatenates the outputs of all backbone CNNs and propagates them through a numbers of layers (1 to 6  layers were evaluated) of linearly decreasing size with a final layer size of 256. This forms the latent vector compatible to the subsequent DeepCAD decoder network.

Unlike the FCN pooling which processes all input views simultaneously, the alternative GRU pooling receives the input views from the backbone CNN sequentially one after the other. This makes it more suitable for varying numbers of images. For evaluation of the GRU pooling stage, we tested different numbers of layers (1 to 8) of identical dimension, different temporal pooling strategies (mean, max, last) and different layer dimensions (64, 128, 256, 512, 1024, 2048). A single fully connected layer is used to achieve the latent vector size of 256. 

Both pooling network variants use rectified linear units (ReLU) as their non-linear activation function in all layers except the last. The final layer generates the latent vector. Here the hyperbolic tanget function ($tanh$) is utilized as it provides output in the range $[-1,1]$ as required for the DeepCAD decoder network.

The final stage of the ARE-Net is formed by the decoder from the DeepCAD library \cite{wu_deepcad_2022} which generates CAD construction sequences from the 256-dimensional latent vector. 

\subsection{Two-stage training}
Training was performed in two stages: First, the full DeepCAD auto-encoder was pre-trained as described in \cite{wu_deepcad_2021}. After this training, the final latent vector of each CAD object from the training set was saved.  Second, simulated image views were rendered from the ground truth CAD sequences and used to train our backbone and multi-view pooling networks. As the loss function, we used the mean-squared error between the predicted latent vectors of the simulated images and the ground-truth latent vectors from the first training stage. We employed the ADAM-optimizer, using 10 epochs during hyper-parameter optimization and 140 epochs for the final model.


\section{Experimental setup}

\subsection{Training data}
\begin{figure}[htb]
  \centering
   \includegraphics[width=0.7\linewidth]{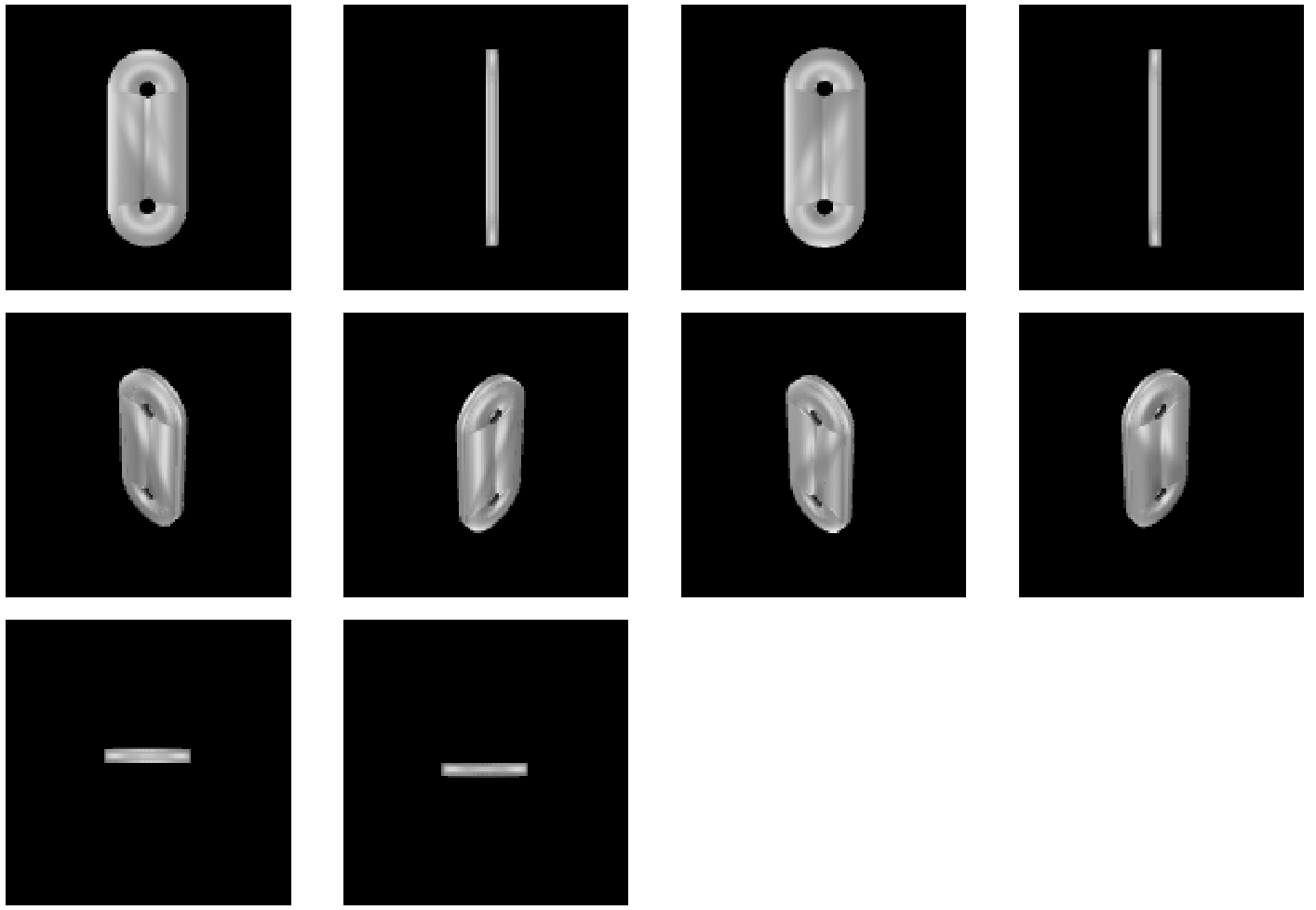}

   \caption{Example of training images from one CAD model: (top row) central view from four sides, (middle row) elevated view, (bottom row) top and bottom views.}
   \label{fig:train_renderings}
\end{figure}

Training images were generated from the DeepCAD dataset consisting of 178,238 CAD models. From each CAD sequence, a 3D mesh object and two different projection datasets were generated: (1) A simple dataset of 128x128 grayscale images from 10 fixed and evenly spaced view angles as shown in \autoref{fig:train_renderings}. (2) A complex dataset of 256x256 RGB images with random but uniform object color from 24 randomly spaced viewing angles. In the second dataset the photogrammetry ground-plane from \cite{boessenecker_coastal_2020} was used as a base on which each model rests. It is composed of non-repeating patterns and is used as a turntable for real objects during the final real world test. The intention is to provide the model with additional information on orientation and scale of the objects, otherwise lost due to the random viewing angles.

For training on the simple dataset, all 10 images were used. When training on the complex dataset, a random selection of 5 to 20 images was chosen. To allow an unbiased comparison of our network to former work by the DeepCAD researchers, the same training-, validation- and testing-split (90\%-5\%-5\%) used in \cite{wu_deepcad_2021} was applied.

\subsection{Hyper-parameter Optimization}
Our model contains several hyper-parameters requiring optimization. General parameters are the learning rate, drop out ratio, weight decay and the number of ResNet backbone layers. The parameters of the two pooling networks are the number and dimensions of layers. For the GRU network, the temporal pooling strategy (mean, max, last) also needed investigation.
In order to identify suitable hyper-parameters such as network attributes and training parameters which remain constant during any given training run an incremental experimentation procedure is followed. For hyper-parameter optimization, the Optuna library \cite{optuna_preferred_networks__inc_optuna_nodate} was used. It allows for efficient search through the high dimensional search space and automatically records results and useful statistics.

\subsection{Accuracy metrics}

To compare the accuracy of the predicted CAD models, three different metrics were employed: 
The command accuracy $ACC_{cmd}$ measures the agreement of the predicted CAD command type $\hat{t}_{i}$ with the ground truth command type $t_{i}$ for a CAD construction sequence of $N_c$ steps:

    \begin{equation}
        ACC_{cmd}=\frac{1}{N_{c}}\sum_{i=1}^{N_{c}}\left( t_{i} == \hat{t}_{i} \right)
        \label{eq:acc_cmd}
    \end{equation}

While $ACC_{cmd}$ measures that fraction of correct commands, the correctness of the continuous parameters of each command shall also be evaluated. The parameter accuracy $ACC_{param}$ quantifies the agreement of a predicted 8-bit CAD parameter $\hat{p}_{i,j}$ to its ground-truth counterpart $p_{i,j}$. Only correctly predicted commands $N_{c2} \leq N_c$ were evaluated and a threshold of $\eta = 3$ was used, as suggested in \cite{wu_deepcad_2021}:

    \begin{equation}
        ACC_{param}=\frac{1}{N_{c2}}\sum_{i=1}^{N_{c2}}\sum_{j=1}^{\left| \hat{p}_{i} \right|}
        \left( \left| p_{i,j}-\hat{p}_{i,j} \right| \leq \eta \right)
        \label{eq:acc_param}
    \end{equation}

For geometric evaluation of the 3D model, the so-called Chamfer Distance $CD$ was used \cite{noauthor_chamfer_nodate,agarwal_learning_2019}. It computes the shortest distance of one point $x$ on the surface $S_1$ of the predicted object to the closest point $y$ on the surface $S_2$ of the ground-truth object. This is carried out in both directions. In this work, 2000 surface points were evaluated per model.

  \begin{equation}
        CD=\frac{1}{S_1}\sum_{x\in S_1}\min_{y\in S_2}\left| \left| x-y \right| \right|\mathrm{}_{2}^{2}+\frac{1}{S_2}\sum_{x\in S_2}\min_{y\in S_1}\left| \left| y-x \right| \right|\mathrm{}_{2}^{2}
        \label{ep:cd}
    \end{equation}
    
\subsection{Benchmark comparison}
As no other method generating CAD models is known to us, comparison is performed using the following two methods: (1) 
The original DeepCAD auto-encoder is fed with ground-truth CAD-sequences to encode a latent vector and decoded again. The predicted CAD sequence is then evaluated by the accuracy metrics described above. By using loss-less input CAD sequences, this approach represents the ideally achievable results in our comparison and will be referred to as the ``baseline''.

(2) For a more realistic comparison, the PointNet++ encoder \cite{qi2017pointnet} was evaluated as a state-of-the-art method. Point-clouds were sampled from the ground-truth 3D objects. The PointNet++ encoder was used to map the point-clouds into a latent vector and then processed by the Deep-CAD decoder as proposed by \cite{wu_deepcad_2021}.

\subsection{Reconstruction from photographic images}
For an initial assessment of the performance of our method on real world images, two test objects were chosen: a cardboard box representing a very simple case and a camera mount as a more complex example. Both are intentionally of uniform color to match the simulated objects seen during training. The objects were placed on a paper version of the photogrammetry ground plane. Then 20 pictures from varying perspectives were taken by a smartphone camera while changing the inclination angle relative to the ground plane and rotating a turntable underneath the object. The image background behind the objects was then cropped away manually. All pictures were sized down to 224x224 pixels and passed into the Automatic Reverse Engineering Network (ARE-Net) with GRU pooling as trained on the simulated complex dataset.

\section{Results}
The best performing hyper-parameters are summarized in \autoref{tab:hpo_results}. On the simple dataset the GRU with a shallow ResNet18 backbone had sufficient distinguishing power, whereas ResNet34 performed better for the simpler FCN network as well as for the GRU for the complex dataset. Three FC layers were optimal for FCN pooling, but more than one layer didn't increase performance of the GRU pooling stages. As for the GRU-specific parameters, sightly larger networks proved favorable for the complex dataset.

\begin{table}[htb]
\centering
\begin{tabular}{l|lll}
                     Pooling network & FCN        & GRU       & GRU \\ 
                     Dataset & simple  & simple  & complex       \\
                     \hline
                     Learning rate             & $1.3\cdot10^{-4}$   & $4.8\cdot10^{-4}$   & $1.5\cdot10^{-4}$    \\
                      Drop out                  & $4.8\%$          & $16.1\%$         & $17.2\%$          \\
                      Weight decay              & $5.45\cdot10^{-5}$  & $3.18\cdot10^{-6}$  & $4.38\cdot10^{-6}$   \\
                      Backbone                  & ResNet34         & ResNet18         & ResNet34          \\
                     \hline
                     FC layers    & $3$              & $1$              & $1$               \\
                     GRU pooling                   & -                & $max$            & $last$            \\
                     GRU layers                & -                & $1$              & $2$               \\
                     GRU dimension             & -                & $1024$           & $2048$            \\
\end{tabular}

\hspace{4mm}

\caption{\label{tab:hpo_results}Best hyper-parameters found by our optimization.}
\end{table}

\autoref{tab:final_results} compares the accuracy metrics of our models using the optimized hyper-parameters. 
It stands out that the GRU pooling network trained on the simple dataset achieved the best overall performance. It reaches an $ACC_{cmd}$ of 92.8\%, an $ACC_{param}$ of 78.8\% and a median CD of 1.75$\cdot 10^3$. However, the fraction of 18.4\% of CAD models that could not be constructed is notably worse than for the point cloud encoder. The percentage of invalid CAD topologies is reported as "CAD model invalid". An invalid sequence may occur, for example, if a curve sketch command is not followed by a 3D extrusion. This tends to occur more often for longer command sequences. 

\begin{table*}[htb]
\centering
\begin{tabular}{l|cccc}
Method                   & $ACC_{cmd}\uparrow$ & $ACC_{param}\uparrow$ & $median\ CD\downarrow$ & $CAD\ model\ invalid \downarrow$  \\ \hline
ARE-Net  FC (simple data)      & 92.14\%      & 74.2\%       & 4.21$\cdot{}10^3$     & 18.1\%            \\
ARE-Net  GRU  (simple data)     & {\bf92.83\%}  & {\bf78.8\%}  & {\bf{1.75$\cdot{}10^3$}}   & 18.4\%            \\
ARE-Net  GRU (complex data)    & 92.78\%       & 74.6\%       & 4.07$\cdot{}10^3$        & 18.8\%            \\
DeepCAD PointNet++       & 84.95\%       & 74.2\%       & 10.3$\cdot{}10^3$       & {\bf12.1\%}       \\ \hline
Baseline: & & & & \\
 DeepCAD auto-encoder      & 99.50\%       & 98.0\%       & 0.75$\cdot{}10^3$       & 2.7\%         \\
\end{tabular}
\hspace{4mm}
\caption[Quantitative results of CAD reconstruction]{Quantitative results of CAD reconstruction of the presented ARE-Net, DeepCAD with point cloud network and the DeepCAD auto-encoder.}
\label{tab:final_results}
\end{table*}

The ARE-Net models trained on the simple datasets surpass the one trained on the complex data. The random variation of perspectives and number of input images during training represent a harder problem which did not provide an advantage in this comparison.

\begin{figure}[htb]
\centering
\includegraphics[width=0.5\linewidth]{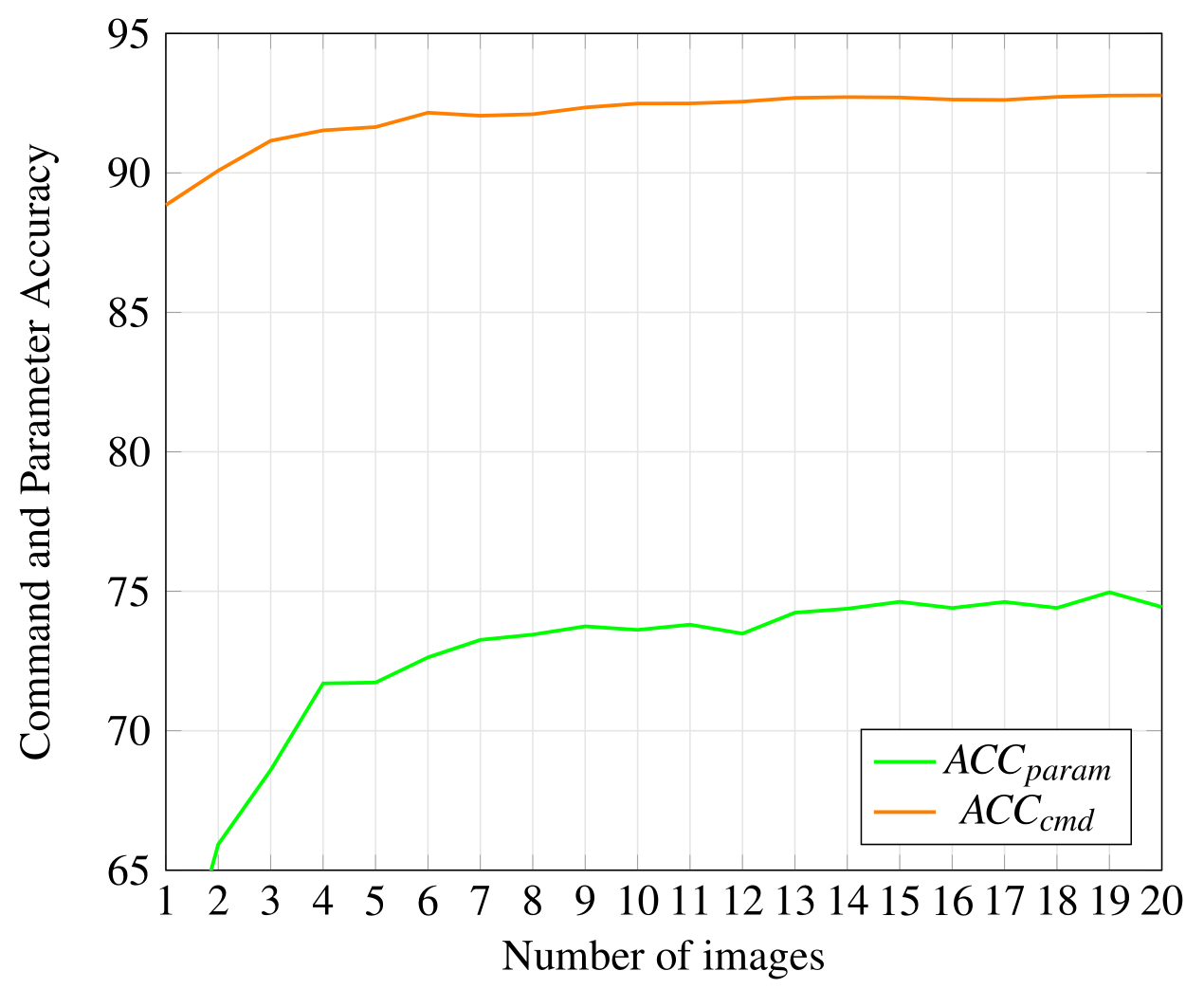}
\caption{Accuracy results for different numbers of input images passed into the ARE-Net using the complete object test set.}
\label{fig:var_num}
\end{figure}

The accuracy on the test set of the ARE-Net with GRU pooling is plotted in \autoref{fig:var_num} as a function of the number of input images. Above 13 images the accuracy barely increases, which is in line with \cite{su_multi-view_2015} describing 12 images as a useful lower bound, beyond which the accuracy of their network levels. 

\begin{figure}[htb]
\centering
\includegraphics[width=0.49\linewidth, height=0.35\linewidth]{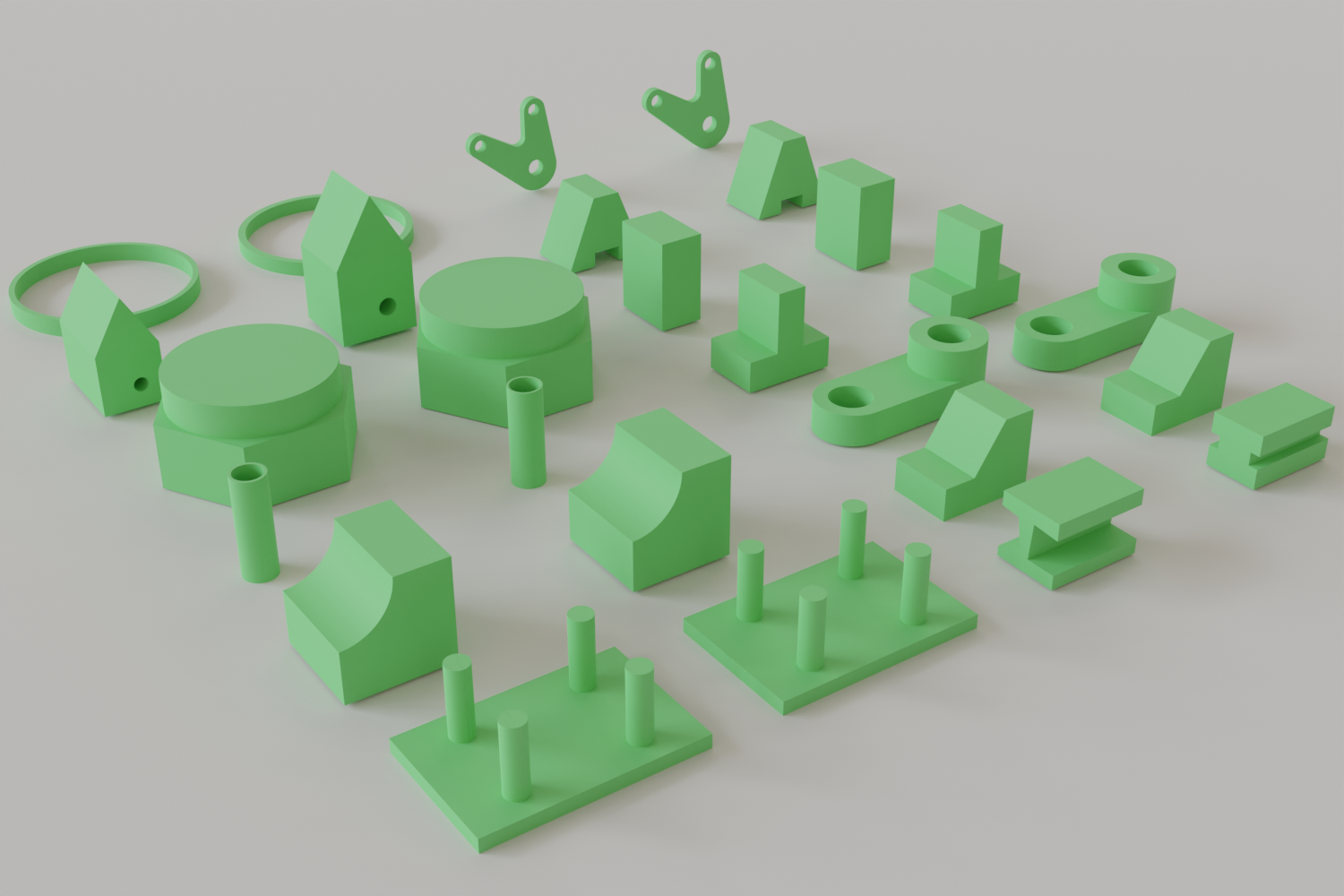}
\includegraphics[width=0.49\linewidth, height=0.35\linewidth]{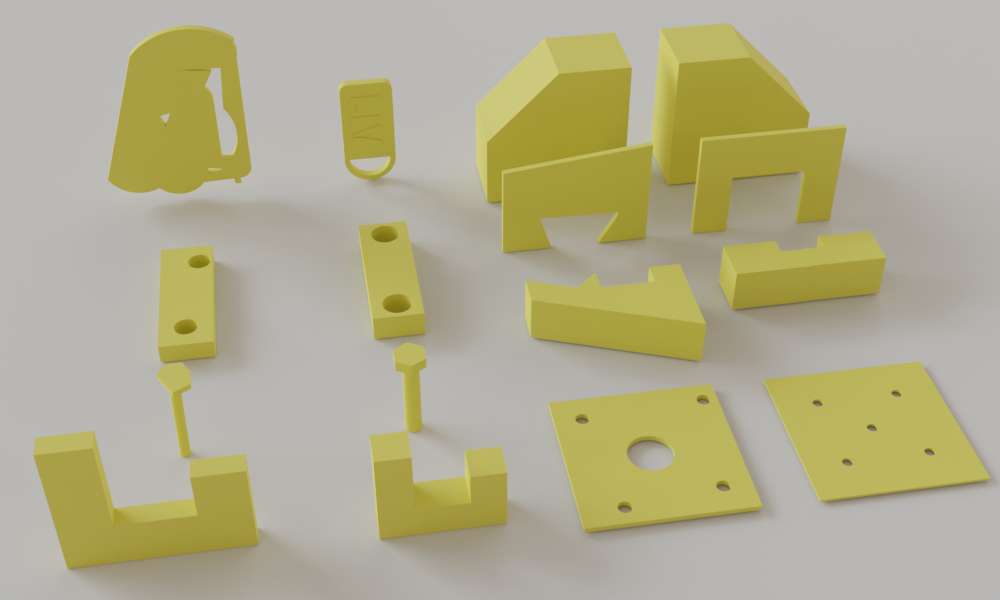}
\caption{Random selection from the test set of representative good (green) and poor (yellow) reconstruction results. The model predictions are shown on the left, next to their corresponding ground-truth models. 
}
\label{fig:res-green-yellow}
\end{figure}

\autoref{fig:res-green-yellow} compares the reconstructed geometries. The following observations can be made:
A variety of reconstructions is quite successful. Often times the network seems to "comprehend" the basic form of the shape present, but lacks the ability to exactly reproduce it quantitatively. For example, regarding the yellow object in the bottom right corner of \autoref{fig:res-green-yellow}, it is clear that the model has recognized the basic shape of the plate and manages to reproduce it quite well. It also extracts the correct number of holes but still fails to reproduce their size and exact position.

\begin{figure}[htb]
\centering
\includegraphics[width=0.49\linewidth, height=0.35\linewidth]{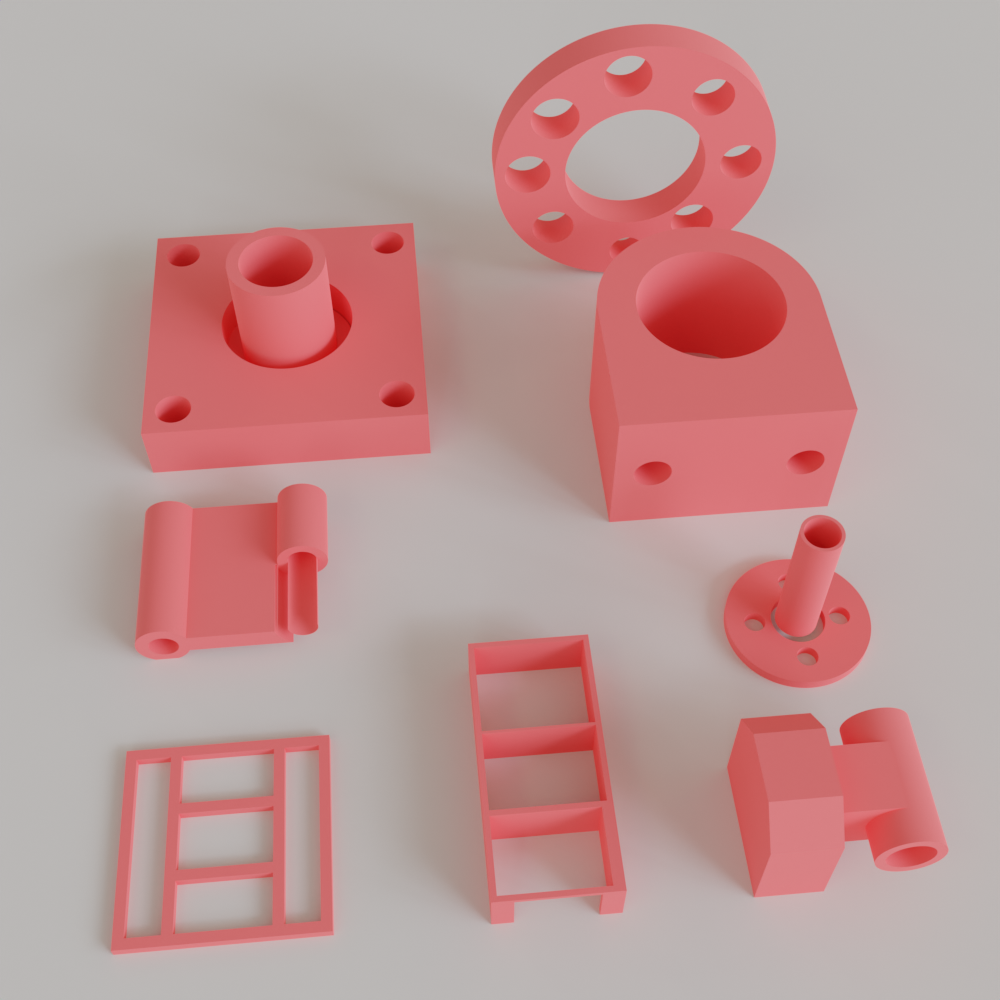}
\caption{Random selection from the test set of ground-truth models that could not be successfully reconstructed.
}
\label{fig:res-red}
\end{figure}

Conversely, a fraction of about 18\% of more complex ground-truth models could not be successfully reconstructed, some examples are show in \autoref{fig:res-red}. Visual comparison shows that these models are generally more complex than their valid counterparts, e.g.~containing holes of different diameters or extrusion into different spatial directions. 

\begin{figure}[htb]
\centering
\includegraphics[width=0.35\linewidth]{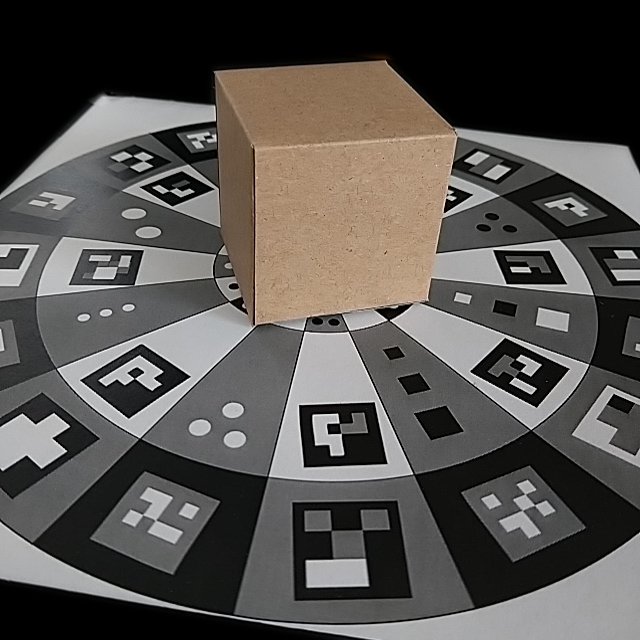}
\hspace{3mm}
\includegraphics[width=0.35\linewidth]{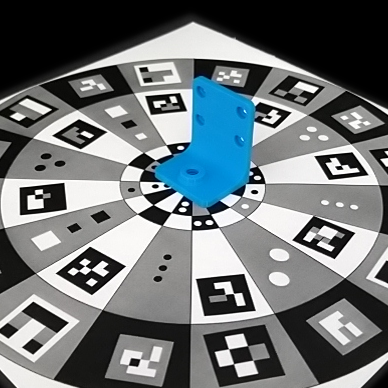} \\
\includegraphics[width=0.35\linewidth]{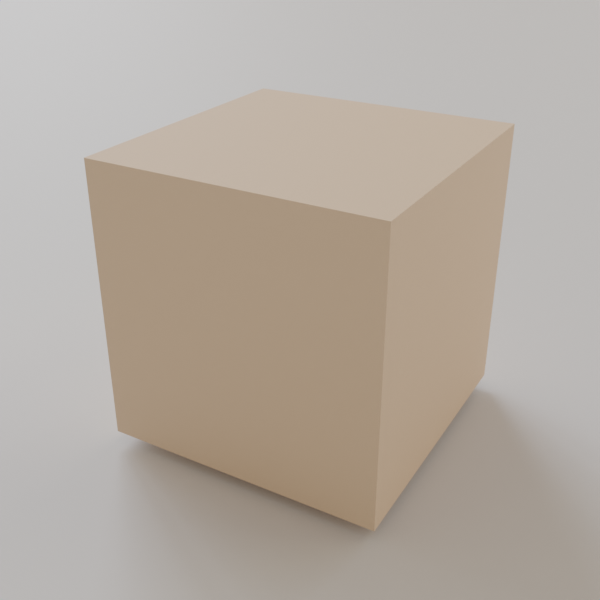}
\hspace{3mm}
\includegraphics[width=0.35\linewidth]{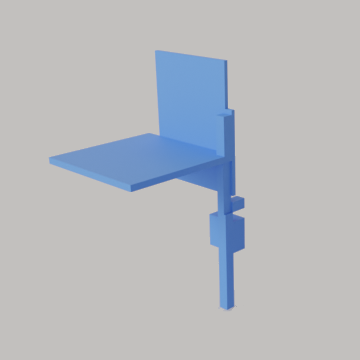}
\caption{Real object reconstruction attempts: The top row show selected photos of the two objects placed on the photogrammetry ground-plane (left: cardboard box, right: camera mount angle). The bottom row shows the respective CAD reconstructions.}
\label{fig:res-real}
\end{figure}

Two representative photos of our real world objects and their reconstructions are shown in \autoref{fig:res-real}. The reconstructed CAD sequence of the cardbox is a perfect cube with equal side lengths, up to the 8-bit precision. 
As for the more complicated camera mount, a valid CAD model could be created from the photos. However, 
only the basic L-shape is represented by the model. The relative dimensions are inaccurate and details like the screw holes are completely missing. Moreover, the reconstruction exhibits a prominent elongated bar at the bottom which is not at all present in the original model.
This second real-world reconstruction was hence only partially successful. 

\section{Discussion and conclusions}

We developed a novel method for end-to-end generation of CAD sequences directly from photographic images using an encoder-decoder network architecture. 
Models were trained in a two-stage approach on 2D renderings of simulated CAD objects and positively evaluated. A first proof-of-concept of the method on real photos was realized.

Two different multi-view pooling stages were compared: a feed-forward fully-con\-nect\-ed network (FCN) and a gated recurrent unit (GRU). A number of hyper-parameters were extensively optimized. Our results show that the additional complexity introduced by the GRU pays off by producing a significant improvement in all three accuracy metrics. Moreover, the GRU takes in the individual images one after the other such that the number of input images can be handled more flexibly. Our experiments confirm the earlier finding~\cite{su_multi-view_2015} that around 12 different views of an object can be considered a practical lower bound, with little improvement above that number. 

Comparing our CAD models reconstructed from rendered images of the test set to reconstructions from 3D point-clouds by the state-of-the art PointNet++ encoder, our encoders successfully created valid CAD sequences in more than 80\% of the cases which is lower than the success rate of the point-cloud encoder. Regarding the accuracy measures, our encoders outperformed the point-cloud encoder by a large margin.

Most importantly, our work establishes the basic feasibility of image-based reverse engineering of 3D CAD models by neural networks. In future applications this might reduce the amount of time-consuming work of highly trained engineers or enable untrained laymen to work with CAD technologies for 3D printing previously inaccessible without specialized training. 

Current limitations of the approach include that the length of CAD sequences is still limited to 60 commands, hence only supporting relatively simple objects. Also our representation is limited to planar and cylindrical surfaces, while many real-world objects may include more flexible triangle meshes or spline representations. 

Furthermore, the exact position and size of object details - especially small holes - must be improved for practical applications. The loss function used to train the DeepCAD decoder network penalizes deviations of the CAD parameters but does not contain a distance metric \cite{wu_deepcad_2021}. We believe that an end-to-end training of the complete model may improve these results, allowing for more specialized loss functions to get a more direct handle on the quantitative sequence parameters.

Future work should also focus on improving the image rendering of the training data. This may include physics-based rendering techniques such as ray-tracing to better simulate real-world cases and the incorporation of reflections, image blur and noise to better mimic an actual picture taken by the end-user. Data augmentation by different backgrounds and model textures should also be considered. Just like the camera view angles, the distance and translation of the object should also be varied. A fine-tuning of the model parameters training with a (limited) set of real-world photos of 3D-printed objects from given CAD models could also be pursued.
Finally different backbone and/or pooling architectures, such as attention based techniques could be explored going forward.

Generally the direction proposed in this work seems promising. It will be interesting to see what this or similar approaches will lead to down the line. One may predict that experts and consumers might soon be using parametric, CAD generating 3D-scanning-applications, just as naturally as optical character recognition (OCR) is used today, saving countless hours of repetitive work and providing unpreceded possibilities of interaction and creation in this three-dimensional world. 

\section*{Acknowledgements}
We would like to thank Rundi Wu and his co-workers for openly sharing their ground-braking DeepCAD work and providing extensive support materials such as their dataset, the generative CAD decoder, the point-cloud encoder and evaluation metrics.

%
%
%
%
\bibliographystyle{splncs04}
\bibliography{086-main.bib}

\end{document}